\documentclass{article}
\usepackage{ijcai26}

\usepackage{times}
\usepackage{soul}
\usepackage{url}
\usepackage[english]{babel}
\usepackage[hidelinks]{hyperref}
\usepackage[utf8]{inputenc}
\usepackage[small]{caption}
\usepackage{graphicx}
\usepackage{subcaption}
\usepackage{amsmath}
\usepackage{amsthm}
\usepackage{booktabs}
\usepackage{algorithm}
\usepackage{algorithmic}
\usepackage[switch]{lineno}
\usepackage{enumerate}
\usepackage{todonotes}
\usepackage{tabularx}
\usepackage{microtype}

\newcommand{\Alphabet}{\mathcal{V}}
\newcommand{\Variables}{\mathcal{X}}

\newcommand{\naf}{\mi{not}\,}

\newcommand{\atm}{\ensuremath{\alpha}}
\def\AS{\mathit{AS}}
\def\mi#1{\mathit{#1\/}}
\newcommand{\Lits}{\mathcal{A}}
\newcommand{\HU}{\ensuremath{\mathit{H\!U}}}
\newcommand{\HB}{\ensuremath{\mathit{H\!B}}}

\newcommand{\nop}[1]{}

\newcommand{\citeNBYB}[1]{\citeauthor{#1}~\shortcite{#1}}

\newcounter{myenumctr}

\newenvironment{myitemize}
{\begin{list}{--}{\setlength{\topsep}{0pt}
\setlength{\leftmargin}{0pt}
\setlength{\itemsep}{0pt}
\setlength{\itemindent}{13.5pt}}}
{\end{list}}

\newtheorem{example}{Example}

\newif\ifrevised\revisedfalse
\revisedtrue
\ifrevised

\newcommand{\oldversion}[1]{*** {\color{gray}#1} ***}

\else

\newcommand{\oldversion}[1]{}

\fi

\title{An XAI View on Explainable ASP: Methods, Systems, and Perspectives}

\author{
Thomas Eiter
\and
Tobias Geibinger
\And
Zeynep G. Saribatur 
\affiliations
Institute of Logic and Computation, TU Wien, Austria
\emails
\{thomas.eiter, tobias.geibinger, zeynep.saribatur\}@tuwien.ac.at
}

\begin{document}

\maketitle

\begin{abstract}
Answer Set Programming (ASP) is a popular declarative reasoning and problem solving approach in symbolic AI.
Its rule-based formalism makes it inherently attractive for explainable and interpretive reasoning, which is gaining importance with the surge of Explainable AI (XAI). A number of explanation approaches and tools for ASP have been developed, which often tackle specific explanatory settings and may not cover all scenarios that ASP users encounter.
In this survey, we provide, guided by an XAI perspective, an overview of types of ASP explanations in connection with user questions for explanation, and describe their coverage by current theory and tools. 
Furthermore, we pinpoint gaps in existing ASP explanations approaches and identify research directions for future work.
\end{abstract}
\section{Introduction}


Understanding the decision-making of AI systems is crucial for 
accepting them as tools in industry and society to ease our everyday life. This need especially occurs for AI systems based on machine learning, as they usually provide little-to-no insights on their underlying models built from observed data.
The field of Explainable AI (XAI) tackles the challenge of AI systems providing explanations on their decisions and their models, to help in the understanding of their mechanism~\cite{calegari2020integration,schwalbe2024comprehensive}. 
Symbolic AI, which is based on 
describing the knowledge of the world and a problem at hand through logical or related formal languages, is in a strong position for XAI as symbolic and especially rule-based representations are more understandable to humans than what subsymbolic AI systems  (largely neural networks) provide, and thus allow for transparency.  


One of the core logic-based formalisms in symbolic AI is Answer Set Programming (ASP)~\cite{aspglance11,DBLP:books/sp/Lifschitz19}, which is based on logic programs under stable model semantics. 
In ASP, a problem is represented by a set of rules of the form $H\leftarrow B$ in a first-order language (called "program"),  where intuitively, $H$ must hold  whenever $B$ is true. 
The models of such programs (called answer sets \cite{gelfond1991classical}) then correspond to the solutions of the problem. The representation often follows a ``guess-and-check'' methodology, where solution candidates are generated (usually via choice rules) and invalid candidates ruled out through constraints.
%
%
A notable feature of ASP is nonmonotonicity, i.e., adding rules to a program may 
cause new answer sets and 
that inferences from all answer sets no longer hold. It embraces the {\em closed world assumption}\/ and can be fruitfully exploited to model exceptions and defaults, e.g., inertia rules in planning. 

Thanks to its declarative and expressive language, the rich support for features such as aggregates, optimisation, and preference handling, and by the availability of efficient solvers, ASP is highly popular and widely used beyond AI in Computer Science and other fields.
It has been used to solve a variety of problems such as combinatorial optimisation, logical reasoning, planning,
bioinformatics, and data integration to name a few \cite{schaub2018special}, with growing use in 
industry \cite{rajaratnam2023solving}. Moreover, recent work shows the potential of ASP for neuro-symbolic 
AI 
\cite{rader25survey,shakarian2023neuro}
and aiding Large Language Models in reasoning \cite{kareem2024using,DBLP:journals/corr/abs-2302-03780}. 

Given that an ASP program is a declarative specification without explicit control flow, users might have issues following the underlying reasoning mechanisms, i.e., why a certain result was returned by the solver.
Several tools and theories for explainability in ASP exist and continue to be investigated. 
Those explanation concepts not only differ in what kind of answer they provide, but also in what kind of setting they operate in. 
A previous landmark survey on explanations in ASP \cite{DBLP:journals/tplp/FandinnoS19} focuses on the questions of why a certain atom is (not) in a given answer set and why no answer set can be computed. However, from an XAI perspective  scenarios, where an ASP user finds themselves with a question about the behaviour of the ASP system, are more diverse than these plain settings.


In this survey, we 
provide an overview of the types of explanations that may arise in ASP and show how they are covered by existing ASP explanation approaches. Specifically:

\begin{myitemize}
 \item 
 By taking an XAI perspective, we structure our survey along Local vs.\ Global explanations, distinguishing explanations for computed answer sets and explanations over the general behaviour of the given program. 
 \item We identify a comprehensive list of typical user questions. While some of them can be natively addressed by existing methods, some cannot. Thus, we show how they can be reduced to other questions where explanations are possible. %

\nop{
While explanations in ASP was the topic of a previous survey~\cite{DBLP:journals/tplp/FandinnoS19}, the authors only considered two contexts of explanations in ASP: 
(1) the user obtained an answer set and wants to inquire why an atom is (not) contained in it, and 
(2) they obtained no answer set at all. 
In addition to these questions, we investigate further questions and how they are addressed. Although some considered questions are not directly addressed in the literature, we discuss how to transform the questions into ones that can be handled by the existing tools.
%
}
\item Naturally, we cover approaches and tools that were not 
available for \citeauthor{DBLP:journals/tplp/FandinnoS19}'s survey. 

\item 
The user-centric view also allows us to highlight explanatory settings where current concepts and tools are lacking, even though those scenarios may be encountered by ASP users. We thus propose 
directions for future research to fill these gaps, as well as to address recent developments in AI. 
\end{myitemize}

This survey is intended to serve both researchers, and ASP users who are not experts, but are looking for explanation tools and methods which fit their 
situation and purpose.


\section{Answer Set Programming}\label{sec:asp}




\paragraph{Syntax.} A \emph{(disjunctive) logic program} is a finite set of \emph{rules}:
\begin{equation*}
\atm_1 \lor \dots \lor \alpha_k \leftarrow \atm_{k+1},\dots,\atm_m, \mi{not}\ \atm_{m+1},\dots,\mi{not}\ \atm_n 
\end{equation*}
where  each $\atm_i$ is an \emph{atom} of form $p(t_1,\dots,t_n)$, where $p$ is a predicate, and each $t_i$ is either a constant or a variable. A \emph{literal} is 
either a formula $\atm$ (\emph{positive literal}) or $\mi{not}\ \atm$ (\emph{negative literal}), where $\atm$ is an atom.
Intuitively, 
$\mi{not}\ \atm$ is true if $\atm$
cannot be derived using rules, and false otherwise; $\naf$ is
called \emph{default} negation.
For a rule $r$ of form $H(r) \leftarrow B(r)$, we refer to $H(r)$ as the \emph{head} of $r$, and to $B(r)$ as the \emph{body} of $r$. The \emph{positive body atoms}\/ of $r$ are denoted by $B^+(r)=\{\atm_{k+1}, \dots, \atm_m\}$, and the {\em negative
body atoms}\/ by $B^-(r)=\{\atm_{m+1},$ $\dots,\atm_n\}$.
A rule $r$ is \emph{normal} if $k=1$ and a \emph{constraint} if $k=0$; 
%
Furthermore, $r$ is \emph{positive} if $n=m$ 
and a \emph{fact} if $k=n=1$ and it is variable-free. 
A rule is \emph{ground} if
all
its literals are variable-free. 
A program $P$ 
is \emph{normal}, \emph{positive}\ or \emph{ground} if its rules are.
Lastly, we use $\mi{not}\ S = \{ \mi{not}\ s \mid s \in S \}$ for any set of atoms $S$ as a shorthand.


\begin{example}\label{ex:initial}
Suppose the manager of the Bremen Town Musicians\footnote{Inspired by the Grimms' Fairy Tale} relies on the following program $P_1$ about donkeys:
    \begin{align}
        &\mi{sold}(D) \leftarrow \mi{donkey}(D), \mi{old}(D), \mi{not} \ \mi{musician}(D) \label{eq:init_sold_rule} \\
        &\mi{musician}(D) \leftarrow \mi{donkey}(D), \mi{not} \ \mi{sold}(D) \label{eq:init_mus_rule} \\
        &\mi{old}(D) \leftarrow \mi{donkey}(D), \mi{age}(D,A), A > 10 \label{rule:old}\\
        &\mi{donkey}(d) \\
        &\mi{age}(d, 25)
    \end{align}
    The program has three non-ground rules and two facts;
   intuitively, a donkey which is 
   classified as old 
   will be sold unless it is a musician, and a donkey is a musician unless it is sold.
   All rules are normal and $P_1$ is thus a normal program.
    However, one could also consider $P_2$
    resulting from $P_1$ by replacing the first two rules with:
    \begin{align}
        &\mi{sold}(D) \lor \mi{musician}(D) \leftarrow \mi{donkey}(D) \label{dis-donkey}\\
        &\leftarrow \mi{sold}(D), \mi{not} \ \mi{old}(D) \label{sold-constraint}
    \end{align}
    Here (\ref{dis-donkey}) is a disjunctive rule and (\ref{sold-constraint}) a constraint. The choice of becoming a musician is now expressed via disjunction.
\end{example}

\paragraph{Semantics.} 
The answer set semantics is defined via ground programs
that are obtained by grounding as follows. 

The \emph{Herbrand universe} of a program $P$, denoted by $\HU_{\!P}$, is the set of all constant symbols 
appearing in $P$,  
and the \emph{Herbrand base} of $P$, denoted by $\HB_P$, is
the set of all ground atoms constructable with predicates from ${\cal P}$ and terms from $\HU_{\!P}$.
The \emph{ground instances of a rule} $r \in P$, denoted by
$\mathit{grd}(r,P)$, is
the set of rules obtained by replacing all variables in $r$ with
constant symbols in $\HU_{\!P}$ in all possible ways.
The \emph{grounding} of $P$ is then 
$\mathit{grd}(P) =
\bigcup_{r\in P} \mathit{grd}(r,P)$.


An \emph{interpretation} $I$  of a ground program $P$ is a subset of
$\HB_P$. Such $I$ (i) \emph{satisfies} a literal $\ell$, denoted by $I \models \ell$, whenever $\alpha \in I$ if $\ell = \alpha$, and $\alpha \not\in I$ if $\ell = \mi{not} \ \alpha$, 
%
(ii) satisfies a rule $r \in P$   (denoted by $I \models r$) if
$H(r) \cap  I \neq \emptyset$ whenever $B^+(r) \subseteq I$ and $B^-(r) \cap I = \emptyset$, and 
satisfies (is a \emph{model} of) $P$ (denoted by $I \models P$), if $I \models r$ for all $r\in P$. A model $I$ of $P$ is \emph{minimal} if there is no $J\models P$ such that $J \subset I$.

The \emph{Gelfond-Lifschitz} (GL-)\emph{reduct}  $P^I$ of a program $P$ relative to an interpretation 
$I$ is the 
program obtained from $grd(P)$
when each rule $H(r) \leftarrow B^+(r),\mi{not}\,B^-(r)$
(i) with $B^-(r) \cap I \neq \emptyset$ is deleted, and
(ii) is replaced by $H(r) \leftarrow B^+(r)$, otherwise.
Informally, the first step removes the rules where $I$ contradicts some default negated literal, while the second step removes the negative body from all others.%

An interpretation $I$ is an \emph{answer set} of a program $P$, 
if it is the minimal model of the GL-reduct $P^I$. 
By $\mi{AS}(P)$ we denote the set of answer sets of a program $P$.
A program $P$ is \emph{unsatisfiable} (or \emph{inconsistent}) if $\mi{AS}(P)=\emptyset$.


\begin{example}[Ex.~\ref{ex:initial} cont'd]
The Herbrand universe of $P_1$ is $\mi{HU}_P= \{ d, 25 \}$.
For  the interpretation $I_1 = \{ \mi{donkey}(d),$ $\mi{age}(d,25), \mi{old}(d), \mi{sold}(d) \}$, we have (among others) the following rules in the reduct $P_1^{I_1}$:\footnote{Rules 
which are trivially satisfied are not shown.}
\begin{align}
    &\mi{sold}(d) \leftarrow \mi{donkey}(d), \mi{old}(d) \\
    &\mi{old}(d) \leftarrow \mi{donkey}(d), \mi{age}(d,25) \\
    &\mi{donkey}(d) \\
    &\mi{age}(d, 25)
\end{align}    

\noindent $I_1$ is a minimal model of $P_1^{I_1}$ and thus $I_1\in\mi{AS}(P_1)$.

\end{example}

\paragraph{Language extensions.} 
The \emph{ASP-Core-2 Input Language Format}~\cite{DBLP:journals/tplp/CalimeriFGIKKLM20} 
defines several language extensions; 
most are 
syntactic sugar but 
convenient in practice. We give here an informal account of some major constructs.

\emph{Aggregates} are atoms of the form $\#\mathit{aggr} \ E \prec u$ where 
$\#\mathit{aggr}$ is the aggregate type
($\#count$, $\#sum$, $\#min$ or $\#max$),
$E$ is an aggregate set, 
$\prec$ is a comparison operator, and
$u$ is a term; 
they can appear only in rule bodies.
\begin{example}[Ex.~\ref{ex:initial} cont'd]\label{ex:agg}
   Let us add the facts $donkey(e)$, $age(e,11)$ to $P_1$, yielding $P_3$. 
   To express that at most one donkey can be a musician, 
   we can add the constraint
    \begin{align}
        &\leftarrow \#\mi{count}\{D : \mi{musician}(D)\} > 1 \label{eq:count_con}
    \end{align}
    The 
    aggregate 
    is true if 
    the predicate $\mi{musician}$ holds for more than one constant; 
    then the constraint "fires" and prevents
    that the interpretation is an answer set.
\end{example}

\emph{Choice rules} are of the form $l \leq \{ e_1;\ldots; e_n \} \leq u \leftarrow B$ and select between $l$ and $u$ atoms from $e_1,\ldots,e_n$; if $l=u$, we may write "$=u$".
\begin{example}[Ex.~\ref{ex:initial} cont'd]
    
    The selection between $\mi{sold}(D)$ and $\mi{musician}(D)$ in (\ref{dis-donkey}) can be expressed with a choice rule:
    \begin{align}
        &\{\mi{sold}(D) ; \mi{musician}(D) \} = 1 \leftarrow \mi{donkey}(D) 
    \end{align}
    The rule enforces that for each donkey $D$ exactly one of $\mi{sold}(D)$ and $\mi{musician}(D)$ has to be true.
\end{example}


\emph{Weak constraints} are of the form $:\sim B.\  [w, t_1,\ldots , t_m]$, where
$B$ is a set of literals,
the $t_i$ are constants or variables, and
$w$ is the \emph{weight}. They incur a penalty $w$ if 
$B$ is satisfied, i.e., the constraint instance is violated. An answer set is then assigned the sum of  all such penalties;
it  is \emph{optimal}, if it has minimal penalty. 
More elaborated versions have priority levels, which allow to conveniently express model preference.
\begin{example}[Ex.~\ref{ex:agg} cont'd]
Instead of enforcing that only one of the two donkeys in $P_3$ can be a musician by the constraint (\ref{eq:count_con}), we can 
use also a weak constraint:
    \begin{align}
        &:\sim \mi{musician}(D) \ [1,D] \label{eq:weak}
    \end{align}
Each ground instance of
(\ref{eq:weak})
violated in an answer set incurs  penalty~1. Hence, any optimal answer set will neither contain $\mi{musician}(d)$ nor $\mi{musician}(e)$ as both can be sold.
\end{example}
Further language extensions such as strong negation, conditional literals etc.\ ease problem encoding.

\section{Forms of Explanations in ASP}


\nop{****
In order to better understand the differences of the many explanation approaches and tools for ASP, we will group them into two broad categories: local and global.
This taxonomy is taken from the XAI field, where local explanations arise whenever a classifier is questioned regarding a particular prediction it made~\cite{adadi2018peeking,DBLP:conf/rweb/Silva22}. 
In difference, a global explanation is sought out to explain the classifier as a whole.

In our setting, we do not have classifiers but programs. However, the local vs. global distinction can be adopted as follows.
Whenever, the user has already been presented with an answer set and their query regards that answer set, we call that a local explanation.

In contrast, user might also ask questions without having an answer set at hand and thus are in a sense asking about general program behaviour. 
This is a global explanation.
***}

Explainable AI (XAI) strives for making the behaviour of AI models, in
particular of neural models such as classifiers, understandable
to the user. While explainability and interpretability are often used
interchangeably~\cite{adadi2018peeking} and confused, the subject is
usually either the output for a specific input, 
or insight into the model as a whole.  This leads to
{\em (local) explanations} (concerning a single data point) and {\em
  global explanations} (all data points), respectively, which in a
logic-based setting can be defined using abduction principles
\cite{DBLP:conf/rweb/Silva22}, linking the notions to minimal
satisfying assumptions and prime implicates. 

We can align explainability of ASP with this distinction as
follows. ASP programs are AI models, and answer sets
of ASP programs are data points. Then {\em local explanations}\/ refer to
the case where the user has 
an answer set
that they ask questions about. In contrast, {\em global explanations}
address user questions about answer sets without having one at hand;
here, the general program behaviour is of concern.

In the following, we will also speak of \emph{justification}, which is sometimes used synonymously with the term \emph{explanation}, and for our purposes we consider a justification a type of explanation.

\subsection{Local Explanations}

Local explanations in terms of ASP amount to explaining the reasoning behind an obtained answer set.
In particular, given an answer set $I$ of a program $P$, a user might ask questions about the presence or absence of certain atoms in $I$.
\medskip

\noindent \emph{\bf $Q_1:$ Given $I\in \mi{AS}(P)$, why $I \models \ell$ for some literal $\ell$?}

\noindent 
These forms of questions are the most common explanation problems with several approaches addressing such questions.
%
%
Here we will briefly summarize the concepts, categorizing each distinct approach as a specific method, denoted by \textbf{(M1)}, \textbf{(M2)}, etc. We start with those which were already covered by the previous survey of \citeNBYB{DBLP:journals/tplp/FandinnoS19}. For more details we refer to that work.

\paragraph{Justifications.}
\emph{Offline justifications} \hypertarget{m1}{\textbf{(M1)}} ~\cite{pontelli2009justifications} are labelled directed graphs which, intuitively, can be seen as a step-by-step derivation of the truth value of the respective atom using the rules of the program, originally defined for ground, i.e., variable free, normal logic programs.  

\emph{ABA-based justifications}  \hypertarget{m2}{\textbf{(M2)}}~\cite{DBLP:journals/tplp/0001T16} 
have their roots in \emph{abstract argumentation}~\cite{dung95}. More specifically, they use arguments to describe the semantics of logic programs and the justifications are derived from trees of conflicting arguments (\emph{attack trees}). A labeled assumption-based argumentation (ABA)-based answer set (LABAS) justification is a graph structure where the vertices are literals and edges are either attacks (representing conflicts between
arguments) or supports (representing derivation relationships).

\emph{Causal justifications} \hypertarget{m3}{\textbf{(M3)}}~\cite{DBLP:journals/tplp/CabalarF17} provide explanations for literals in a given answer set based
on a causal semantics for logic programs. The basis for causal
justifications are algebraic terms representing joint causation
through multiplication and alternative causes through addition.
These causal terms can be presented as directed graphs showing
the causal dependencies between literals.



\emph{Justifications in rule-based ASP computation}  \hypertarget{m4}{\textbf{(M4)}}~\cite{beatrix16} stem from the way the ASP solver ASPeRiX computes answer sets. 
Unlike to most solvers, it operates on the non-ground program and non-deterministically chooses rules rather than atoms.
During this computation, the truth value of atoms is iteratively established and the \emph{reason} for it being in the answer set is a set of ground rules which intuitively can be used to derive the atom in question.

Now we come to more recent methods or extensions.

\paragraph{Extended Justifications.}
The 
notion of offline justification has been recently extended to handle 
choice rules and constraints~\cite{trieu2021} 
and 
implemented in the xASP2 system, also supporting aggregates~\cite{DBLP:journals/logcom/AlvianoTSB24}.

One of the main changes in xASP2 is that it computes the 
\emph{well-founded model} of the given program $P$ first.\footnote{Briefly, the well-founded model $\mathit{WFM}(P)$ is a 3-valued approximation of
$\AS(P)$ where atoms true (resp.\ false) in $\mathit{WFM}(P)$ are true (resp.\ false) in each
$I \in \AS(P)$ \cite{van1991well}. 
}
The explanation identifies which atoms are false in the well-founded model
and determines a minimal set of additional atoms that must
be assumed false. Together with the program facts, these
allow deriving the atoms in $I$ through rule application.
%
The overall explanation is constructed in
two phases: first, program rewriting and ASP meta encodings
compute the minimal assumption set; second, these assumptions are used to generate a tree-like justification graph.
The tree-like graph represents the rule application sequence yielding
the target literal $\ell$, where internal nodes correspond to rule
applications and leaves are either facts, atoms false in the
well-founded model, or assumptions.


\begin{example}[Ex.~\ref{ex:initial} cont'd]



    \begin{figure}
        \begin{subfigure}{{0.5\columnwidth}}
        \centering
        \includegraphics[scale=1]{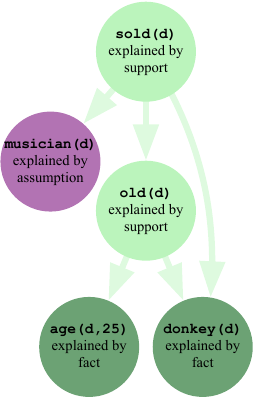} 
        \caption{xASP2 explanation graph}
         \label{fig:xasp2}
        \end{subfigure}~
         \begin{subfigure}{{0.5\columnwidth}}
        \noindent{\small
    \verb;   *; \\
    \verb;   |__sold(d); \\
    \verb;   |  |__donkey(d); \\
    \verb;   |  |__old(d); \\
    \verb;   |  |  |__donkey(d); \\
    \verb;   |  |  |__age(d,25);
\bigskip\bigskip\bigskip}
        \caption{\texttt{xclingo} support graph}
        \label{fig:xclingo}
        \end{subfigure}
        \caption{Explanation for $\mi{sold}(d)$ in 
        $I \in \mi{AS}(P_1)$
        }
        \label{fig:exp}
    \end{figure}
    When we use xASP2 to explain why $\mi{sold}(d)$ is in $I$, 
    its browser displays the 
    graph in Fig.~\ref{fig:xasp2}.\footnote{For space reasons, Fig.~\ref{fig:xasp2} shows a recreation.}
    %
The (green) root node $\mi{sold}(d)$ is connected via
rule (\ref{eq:init_sold_rule}) to its positive body literals $\mi{donkey}(d)$  and $\mi{old}(d)$,
which are also true in $I$ and displayed in green. The negative
body literal $\mi{musician}(d)$  is false in $I$; this is represented as
a purple assumption node, indicating it must be assumed
false for the derivation to hold. The atom $\mi{old}(d)$ (middle
node) is further linked to its supporting facts $\mi{age}(d,25)$
and $\mi{donkey}(d)$ dark green leaf nodes) via rule (\ref{rule:old}). This
tree-like structure shows how $\mi{sold}(d)$ is derived from the
program facts together with the assumption that $\mi{musician}(d)$ 
is false.
\end{example}


Another novel approach is explanations via \emph{unit-provable unsatisfiable subsets (1-PUS)}  \hypertarget{m5}{\textbf{(M5)}},  which are the basis of the UCOREXPLAIN system~\cite{DBLP:conf/lpnmr/AlvianoHSW24}.
%
The concept of a 1-PUS is similar to 
the notion of minimal unsatisfiable sets ~\cite{DBLP:journals/ai/AlvianoDFPR23} 
which is defined relative to a given set  $O$ of \emph{objective atoms},
and is a minimal subset $U\subseteq O$ such that $P \cup \{ \{o\}\geq 0 \leftarrow \mid o \in O \} \cup \{  \leftarrow \mi{not} \ u \mid u \in U \}$ has no answer sets.
A 1-PUS differs by not 
checking the unsatisfiability of the amended program; instead, a simple set of (incomplete) derivation rules is used to check whether we can derive a contradiction. Furthermore, subset minimality is replaced with a user specified lexicographic order. 
In UCOREXPLAIN, the objective atoms $O$ are auxiliary atoms that are injected into the rule bodies in the input program $P$. Intuitively, $o\,{\in}\, O$ controls whether a particular rule
is enabled. Furthermore, the system adds a special rule to the program which makes it inconsistent whenever the literal $\ell$ we seek to explain is true. 
Hence, a 1-PUS in this setting represents a preferred set of rules which, in conjunction with simple derivation rules, can be used to derive $\ell$.

The justifications provided by UCOREXPLAIN can then be displayed in a graph similar to xASP2. An important difference being that in xASP2, an atom $a$ can only be justified by a rule $r$ if $a \in H(r)$, which is not so in UCOREXPLAIN.


\emph{Explanation trees} are obtained from a 
class of a vertex-labeled tree of a program, 
with variations such as shortest and $k$ different explanations
~\cite{erdem2015generating}.

\paragraph{Support Graphs.}

 \emph{Support graphs}  \hypertarget{m6}{\textbf{(M6)}} are defined over labelled programs, where each rule $r$ is given a label $\mathit{Lb}(r)$~\cite{DBLP:journals/tplp/CabalarM24}. 
Furthermore, $\mathit{LB}(P) = \{ \mathit{Lb}(r) \mid r \in P \}$ is the set of all labels in program $P$.
For each atom $p$ in an answer set $I$, they then define the notion of \emph{support} which is defined as $\mathit{SUP}(P,I,p) = \{ r \in P \mid p \in H(r), I\models B(r) \}$, i.e., the set of rules of $P$ which have $p$ in their head and have a true body w.r.t. $I$.
Now, a support graph is a labelled directed graph $G= (I,E,\lambda)$ whose nodes are the atoms of the answer set $I$, edges $e \subseteq I \times I$ connect atoms, and $\lambda : I \rightarrow Lb(P)$ is an injective function assigning a label to each atom.
Furthermore, it is required that for every $p \in I$, the rule $r$ for which $\lambda(p) = \mathit{Lb}(r)$ satisfies $r \in \mathit{SUP}(P,I,p)$ and $B^+(r)= \{ q \mid (q,p) \in E \}$. 
Intuitively, $p$ is connected to all positive body atoms of $r$.
If a support graph $G$ is acyclic, it is called an \emph{explanation}.
The system \texttt{xclingo} uses support graphs to provide explanations~\cite{DBLP:journals/tplp/CabalarM24}. It can handle choice rules, but due to the nature of support graphs, it can only give explanations for atom which are in the given answer set.

\begin{example}[Ex.~\ref{ex:initial} cont'd]
\nop{
    Reconsider once more program $P_1$ and the answer set $I = \{ \mi{donkey}(d),$ $\mi{age}(d,25)$, $\mi{old}(d)$, $\mi{sold}(d) \}$.
    To explain $\mi{sold}(d)\in I$, \texttt{xclingo} outputs

\noindent{\small
    \verb;    *; \\
    \verb;    |__sold(d); \\
    \verb;    |  |__donkey(d); \\
    \verb;    |  |__old(d); \\
    \verb;    |  |  |__donkey(d); \\
    \verb;    |  |  |__age(d,25);
}\smallskip 
}
Figure~\ref{fig:xclingo} shows the output of \texttt{xclingo} to explain $\mi{sold}(d)\in I$.
The atom $\mi{sold}(d)$ is connected to $\mi{donkey}(d)$ and $\mi{old}(d)$
to derive it  using rule (\ref{eq:init_sold_rule}).

\end{example}

Furthermore, \texttt{xclingo} allows the ASP
programmer to attach natural language
messages to rules via annotations (e.g., \%\# message "...").
These messages replace technical rule references in explanations, enabling domain-specific explanations for end-users without ASP expertise.

\paragraph{Witnesses.}
Similarly, \emph{witnesses}  \hypertarget{m7}{\textbf{(M7)}} \cite{wang2022} are solver-independent but aim only at explaining positive literals. The notion aims at explaining why a set $S$ of such
literals is in a given answer  set $I$ of $P$ by stepwise logical derivation, where in each step a set of rules in a reduct of $P$ w.r.t.\ $I$ augmented with already derived atoms is considered; for $S=I$, this amounts to explaining why $I$ is an answer of $P$. For each step, a proof method, in particular resolution, may be chosen. In this way, an explanation of why $I=\{a,b\}$ is an answer set of $P = \{ a\lor
b,\ a \leftarrow b,\ b\leftarrow a\}$, can be given without
resorting to guessing: 
for $S=\{a,b\}$, derive in step 1 $a$ from $a \lor b.$ and $a
\leftarrow b$ by resolution, and in step 2 $b$ from $a$ and
$b\leftarrow a$. Recently, witnesses have been extended to programs with constraint atoms \cite{ijcai2025p523}.

\paragraph{Transform question into unsatisfiability.}
Interestingly, $Q_1$ 
can be transformed into the question of why a program has no answer set.
For example, 
explaining why $a \in I$ 
reduces to explaining why $P\cup \{ \leftarrow b \mid b\not\in I\} \cup \{ \leftarrow \neg b \mid b\in I, b\neq a\} \cup \{ \leftarrow a\}$ 
has no answer sets.
The intuition is that we add the literals true/false in $I$ as assumptions except for $a$ which is forced to be false. 
This idea was 
implicitly applied 
by~\citeNBYB{DBLP:conf/lpnmr/AlvianoHSW24}.
It should be noted that for programs with choice rules the amended program may still be satisfiable. However, in such a case the literal is justified by the choice and there is no better explanation.
Hence, explaining why a program is inconsistent, which is already of great interest by itself, could be a stepping stone to a more universal explanation approach. 


\medskip


\noindent  \emph{\bf $Q_2:$ Given $I\in \mi{AS}(P)$, why $I \models \ell$ but $I \not\models \ell'$?
}

\noindent
\citeNBYB{miller2018explanation} argues that why-questions of form $Q_1$ 
are unequipped for providing proper explanations and instead questions should be seen as contrastive in the sense of ``Why $X$ and not $Y$?''. In our setting this translates to asking $Q_2$. Although such a question can be transformed into unsatisfiability similar to above, which was considered for (weaker) questions of form ``Why not $Y$?" \cite{bogatarkan2020explanation}, dedicated explanation techniques exist in ASP. 

\paragraph{Contrastive Explanations for ASP.}
\citeNBYB{DBLP:conf/jelia/EiterGO23} give a formal framework for \emph{contrastive explanation}  \hypertarget{m8}{\textbf{(M8)}} in ASP.
In detail, their approach assumes that the user wants to know why for a set of atoms $E \subseteq I$ is included in the answer set instead of another set of atoms $F\cap I=\emptyset$.
In order to provide an answer to this question, a \emph{counterfactual account} is generated which consists of a program $P'$ with answer set $I'$ such that $E\cap I' = \emptyset$ and $F \subseteq I'$ and $P'$ is ``similar'' to $P$. The latter means that $P'$ retains as many rules as possible from $P$ and the rules $r \in P' \setminus P$ are facts from a given set of assumables. 

A counterfactual explanation is then a triple $(Q_1,Q_2,Q_\Delta)$, where $Q_1$ and $Q_2$ are essentially minimal sets of ground rules that derive $E$ w.r.t.\ $P$ and $I$, and $F$ w.r.t.\ $P'$ and $I'$, respectively, and $Q_\Delta = P \setminus P'$ is the set of rules that were removed.

The contrastive explanation results from the counterfactual explanation by taking the symmetric difference of $Q_1$ and $Q_2$ and removing the (optional) fixed rules, 
which must be in $P'$.

\begin{example}[Ex.~\ref{ex:initial} cont'd]\label{ex:contr}
    Consider again $P_1$ and assume we add the following constraint yielding $P_4$:
    \begin{align}
        &\leftarrow \mi{musician}(D), \mi{age}(D,A), A > 21 
    \end{align}

    Now, the grounding of the program $P_4$ includes the following rules:
    \begin{align}
        &\mi{sold}(d) \leftarrow \mi{donkey}(d), \mi{old}(d), \mi{not} \ \mi{musician}(d) \label{rule:grd_sold} \\ 
        &\mi{musician}(d) \leftarrow \mi{donkey}(d), \mi{not} \ \mi{sold}(d) \label{rule:grd_mus}  \\
        &\mi{old}(d) \leftarrow \mi{donkey}(d), \mi{age}(d,25), 25 > 10 \label{rule:grd_old} \\
        &\leftarrow \mi{musician}(d), \mi{age}(d,25), 25 > 21 \label{eq:constrain_under21}
    \end{align}
    
    It holds that $\mi{AS}(P_4) \,{=}\, \{ I_1 \}$. Suppose a user asks 
    why
    $\mi{sold}(d) \,{\in}\, I_1$ rather than $\mi{musician}(d) \,{\in}\, I_1$, and 
    the facts are fixed.
    The only contrastive explanation is then $(\{ (\ref{rule:grd_sold}), (\ref{rule:grd_old}) \}, \ \{ (\ref{rule:grd_mus}) \}, \ \{ \ref{eq:constrain_under21})$ intuitively meaning that (\ref{rule:grd_sold}) and (\ref{rule:grd_old}) explain why $\mi{sold}(d) \in I_1$, but if (\ref{eq:constrain_under21}) were removed, we could get $\mi{musician}(d) \in I_1$ via (\ref{rule:grd_mus}).
\end{example}


\medskip

\nop{
\noindent  \emph{\bf $Q_3:$ Why is interpretation $I$ not an answer set of $P$?}

This question has
appeared in the context of debugging~\cite{brain2007debugging,gebser2008meta,oetsch2010catching} which we overview in the next section when explaining unsolvability. 

\todo[inline]{not actually local, merge with $Q_6$}
\medskip
}

\noindent  \emph{\bf $Q_3:$ Why is answer set $I$ of $P$ reported as optimal?}

\noindent
For programs with weak constraints, explanations for optimality would be of great interest. Although currently no method is specifically tailored to this question, it is possible to reduce it to checking why there is no answer set that is better. This can be done by introducing a constraint enforcing strictly less penalty than $I$. Explanation methods for the non-existence of answer sets, which we discuss later, can then be employed. 
While we classified this question as local, it could also be seen as global as the only thing we take from $I$ is its cost.

\medskip

\subsection{Global Explanations}
Global explanations aim to explain the model and its logic as a whole \cite{adadi2018peeking}. In terms of ASP, this 
concerns the following questions.

\medskip

\noindent  \emph{\bf $Q_4:$ Why is $a$ (not) in some answer set of $P$?}

\noindent
Explaining why $a$ appears, or does not appear, in some answer set of $P$ can be seen as local where the $I$ is left implicit. Hence, a local explanation, of a computed witness $I$, can be given. However, obtaining global explanations without needing a specific answer set at hand is also possible.

%
 \paragraph{Why-not Provenance.} This question has been tackled by \emph{why-not provenance}  \hypertarget{m9}{\textbf{(M9)}}~\cite{DBLP:conf/lpnmr/DamasioAA13} which provides justifications for the truth values of atoms w.r.t. answer set semantics for normal logic programs. 
This notion has formal similarities with causal justifications, but rather than causal reasons it gives possible modifications of the program that are needed to ensure a certain truth value. 


\paragraph[Top-down Computation]{Top-down Computation \hypertarget{m10}{\textbf{(M10)}}} 
The Constraint Answer Set Solving system \emph{s(CASP)}~\cite{arias18} can provide justification trees for atoms contained in some answer set due to its implementation in Prolog.
The system finds stable models by goal-directed search and top-down evaluation operating on the non-ground program. This means that the user essentially queries for a certain -- potentially negated -- predicate and is given an answer set where a particular ground instance of the predicate is contained.
The process is top-down in the sense that it starts from the queried predicate and looks whether a ground instance can be justified in an answer set. 
This has the side effect of producing a justification tree which can be displayed to the user. 
Furthermore, this explanation also operates on the non-ground program, thus potentially yielding more compact explanations.

\begin{example}[Ex.~\ref{ex:initial} cont'd]
 If we run s(CASP) for the ground query $\mi{sold}(d)$ on $P_1$ and ask for a justification, it outputs
\smallskip

\noindent{\footnotesize
    \verb;'sold' holds (for d), because; \\
    \verb; 'donkey' holds (for d), and; \\
    \verb; 'old' holds (for d), because; \\
    \verb;  'donkey' holds (for d), justified above, ; \\
    \verb;  and 'age' holds (for d, and 25).; \\
    \verb; there is no evidence that 'musician' holds; \\
    \verb; (for d), because; \\
    \verb;  'donkey' holds (for d), justified above, ; \\
    \verb;  and it is assumed that 'sold' holds (for d); 
}
%
%
%
\end{example}

In a similar fashion, the Satoh-Iwayama procedure~\cite{DBLP:conf/elp/SatohI92,DBLP:conf/prima/DauphinS19}, which is also goal-directed, can be used to answer $Q_4$. 
It also produces a derivation tree that can be given as a justification, but
different from s(CASP), the procedure operates on ground programs.


\medskip

\noindent \emph{\bf $Q_5:$ Why is $a$ in no or in every answer set of $P$?}

\noindent
Dedicated explanation concepts related to questions of this form are \emph{proof systems}  \hypertarget{m11}{\textbf{(M11)}} for ASP. 
For example,  
\citeNBYB{bonatti01} introduced a resolution calculus for sceptical entailment, which is when $a$ appears in every answer set, and a tableaux calculus was introduced by \citeNBYB{gebser13}.
Recently, \citeNBYB{DBLP:conf/ijcai/EiterG25} presented a sequent calculus for sceptical entailment in equilibrium logic, which is 
a logical underpinning of ASP.

Similar to the questions in the previous section, answering why an atom $a$ is in no (respectively, every) answer set of $P$ can be reduced to explaining why $P\cup \{ \leftarrow \neg a \}$ (respectively, $P\cup \{ \leftarrow a \}$) has no answer set. 

\medskip

\noindent  \emph{\bf $Q_6:$ Why does $P$ have no answer sets?}

\noindent
Generally, the works studying this question 
concern \emph{debugging}, i.e., they aim not at a user seeking general explanations but rather to provide technical feedback to the ASP engineer who was expecting a different result, or \emph{proof logging}, where solvers provide verifiable certificates for their results. 
\begin{table*}[ht!]
\centering
\small
\begin{tabularx}{\textwidth}{l c l l l l l}
\toprule
\textbf{Method (System)} &
\textbf{Type} &
\textbf{Questions} & \textbf{Prog.} &
\textbf{Pos./Neg.} &
\textbf{Dep.} &
\textbf{Output} \\
\midrule
\hyperlink{m1}{\textbf{(M1)}}  Offline Justifications (xASP)&
L &
$Q_1,Q_4^*$ & Normal &
Both &
No &
Graph \\

\hyperlink{m2}{\textbf{(M2)}}  ABA-based Justifications (LABAS) &
L &
$Q_1,Q_4^*$ & Normal&
Both &
No &
Argumentation graph \\

\hyperlink{m3}{\textbf{(M3)}}  Causal Justifications &
L &
$Q_1,Q_4^*$ & Normal&
Both &
No &
Graph / algebraic term \\

\hyperlink{m4}{\textbf{(M4)}}  Rule-based Justifications &
L &
$Q_1,Q_4^*$ & Normal&
Both &
Yes &
Set of rules \\

\hyperlink{m5}{\textbf{(M5)}}  1-PUS (UCOREXPLAIN) &
L &
$Q_1,Q_4^*$ & Disj.&
Both &
No &
Graph of derivations \\

\hyperlink{m6}{\textbf{(M6)}}  Support Graphs (xclingo) &
L &
$Q_1,Q_4^*$ & Disj.&
Positive  &
Yes &
Graph \\

\hyperlink{m7}{\textbf{(M7)}} Witnesses &
L &
$Q_1,Q_4^*$ & Disj.&
Positive &
No &
Sequence of rule sets\\

\hyperlink{m8}{\textbf{(M8)}}  Contrastive Explanations &
L &
$Q_2$ & Disj. &
Both &
No &
Structured explanation \\

\hyperlink{m9}{\textbf{(M9)}}  Why-not Provenance &
G &
$Q_4$ & Normal &
Both &
No &
Provenance graph \\
\hyperlink{m10}{\textbf{(M10)}}  Top-down Computation (s(CASP)) &
G &
$Q_1^*, Q_4$ & Normal &
Both &
Yes &
Justification tree \\

\hyperlink{m11}{\textbf{(M11)}}  Proof Systems &
G &
$Q_1^*,Q_2^*,Q_3^*,Q_5,Q_6$ & Disj. &
n/a  &
Yes/No &
Proof certificate \\

\hyperlink{m12}{\textbf{(M12)}}  Debugging (spock, Ouroboros) &
G &
$Q_1^*,Q_2^*,Q_3^*,Q_5^*,Q_6$ & Disj. &
n/a  &
No &
Meta-interpretation \\

\hyperlink{m13}{\textbf{(M13)}}  MUS-based Debugging (DWASP) &
G &
$Q_1^*,Q_2^*,Q_3^*,Q_5^*,Q_6$ & Disj. &
n/a  &
No &
Min. unsat. rule sets \\

\hyperlink{m14}{\textbf{(M14)}}  Stepping &
G &
$Q_1^*,Q_2^*,Q_3^*,Q_5^*,Q_6$ & Disj. &
n/a  &
No &
Interactive trace \\

\hyperlink{m15}{\textbf{(M15)}}  Abstraction &
G &
$Q_1^*,Q_2^*,Q_3^*,Q_5^*,Q_6$ & Normal&
n/a &
No &
Abstracted program \\
\bottomrule
\end{tabularx}
\caption{Overview of explanation techniques in Answer Set Programming.}
\label{tab:overview}
\end{table*}

\paragraph{Debugging.}
One avenue of research into debugging is given by \emph{meta-encoding techniques}  \hypertarget{m12}{\textbf{(M12)}} which, intuitively, provide reasons why each interpretation is not an answer set. Systems that implement this approach are spock~\cite{brain2007debugging,gebser2008meta} and Ouroboros~\cite{oetsch2010catching}; for the latter no a-priori knowledge of inconsistency is needed when the user provides an interpretation whereas spock needs no such information.  
Another approach is to remove the inconsistency by utilizing
\emph{minimal unsatisfiable sets (MUS)}  \hypertarget{m13}{\textbf{(M13)}}, as done e.g.\ in DWASP~\cite{DBLP:journals/tplp/DodaroGRRS19}. A minimal unsatisfiable set contains (non-ground) rules of the program which cannot jointly hold, but if one of them were removed, then the program would be consistent. This concept has roots in model-based diagnosis and constraint satisfaction and has been generalized to nonmonotonic logics including ASP~\cite{eiter2014finding,DBLP:journals/ai/BrewkaTU19} taking the nonmonotonicity of the stable model semantics into account.
\emph{Stepping}  \hypertarget{m14}{\textbf{(M14)}}~\cite{DBLP:journals/tplp/OetschPT18} is an interactive debugging approach that lets the user build an intended answer set step by step, illustrating where the process fails. For more details on debugging techniques we refer to the survey of \citeNBYB{DBLP:journals/tplp/FandinnoS19}.

\paragraph{Proof Systems.}
For $Q_5$, we mentioned proof systems for ASP, which can also serve as a basis to answer question $Q_6$, as it is essentially asking why $\bot$ is derivable.


Furthermore, for certifying the result whenever a solver reports ``unsat'', specialised \emph{proof logging} systems for SAT were adapted to ASP \cite{DBLP:journals/tplp/AlvianoDFHPR19,DBLP:conf/kr/ChewCS24}. Those systems output resolution proofs for the SAT clauses generated by the solver to derive $\bot$.


\paragraph{Abstraction.}
In contrast to debugging and proof approaches, \emph{abstraction}  \hypertarget{m15}{\textbf{(M15)}} notions in ASP highlight the reason for unsolvability through omitting or clustering irrelevant details \cite{DBLP:journals/tplp/SaribaturE21,DBLP:journals/ai/SaribaturES21}. 
Here, abstraction is considered as an \emph{over-approximation}, where a program $P'$ over $\Lits'$ is called an \emph{abstraction} of a program $P$ over $\Lits$, if there exists a mapping $m:\Lits \rightarrow \Lits'\cup \{\top\}$ such that for each answer set $I$ of $P$, $I'=\{m(l) \mid l \in I\}$ is an answer set of $P'$, i.e., $m(\AS(P))\subseteq \AS(P')$. \emph{Omission-based abstraction} \cite{DBLP:journals/tplp/SaribaturE21} relies on a mapping $m_A:{\cal A} \rightarrow \Lits \cup \{\top\}$ for a set $A$ of atoms to be omitted, such that $m_A(\alpha)=\top$ if $\alpha \in A$ and $m_A(\alpha)=\alpha$ if $\alpha \in {\cal A}\setminus A$. Thus $m_A(\AS(P))$ amounts to projecting away the omitted atoms, denoted as $\AS(P)|_{\overline{A}}$. An abstract program $\mi{omit}(P,A)$ is constructed by a syntactic operator that removes rules with heads in $A$ and add choice rules to the head of rules with body atoms occurring in $A$. This ensures over-approximation of the (projected) answer sets.
An \emph{(answer set) blocker set}  refers to a set $R$ of atoms that are kept in the vocabulary, while the rest is omitted from a ground program $P$ to obtain 
the program $\mi{omit}(P,\Lits\setminus R)$, 
such that $\mi{omit}(P,\Lits\setminus R)$ is also unsatisfiable; 
the atoms in $R$ are thus blocking occurrence of answer sets. 

\begin{example}[Ex.~\ref{ex:contr} cont'd]
Assume we add the following rules to $P_4$, yielding $P_5$ that is unsatisfiable:
    \begin{align}
        &\leftarrow \mi{not}\ \mi{musician}(D), \mi{canSing}(D)\\
        &\mi{canSing}(d)
    \end{align}
A blocker set for the grounding of the program $P_5$ would be $R=\{donkey(d), musician(d), age(d,25), canSing(d)\}$, with resulting abstraction $\mi{omit}(\mi{grd}(P_5),\Lits\setminus R)$ as below that is still unsatisfiable.
   \begin{align}
        \{\mi{musician}(d)\} &\leftarrow \mi{donkey}(d)\\
        &\leftarrow \mi{musician}(d), \mi{age}(d,25), 25 > 21\\
        & \leftarrow \mi{not}\ \mi{musician}(d), \mi{canSing}(d)\\
        &\mi{canSing}(d)\\
        &\mi{donkey}(d)\\
        &\mi{age}(d,25)
    \end{align}
In fact $R$ would be the minimal blocker set, as omitting further atoms would no longer preserve unsatisfiability.
\end{example}

A CEGAR-style~\cite{DBLP:journals/jacm/ClarkeGJLV03} abstraction and refinement methodology
enables automated computation of these atom sets. 
This is lifted to the non-ground case for clustering of domain elements  \cite{DBLP:journals/ai/SaribaturES21} to find abstractions that ``zoom-in" on the unsatisfiable part of the program. 





\subsection{Summary}

Table~\ref{tab:overview} presents an overview of the explanation techniques we addressed in this section; 
the symbol $^*$ means one can reduce the question to another one and use the latter's method, like how we can explain why an atom is in some answer set ($Q_4$) by providing a witness and using it for the explanation ($Q_1$). We show which questions these techniques can explain, which program type they support, whether they support explaining positive or negative atoms (or both), if they are solver-dependent, and their explanation output. We can observe that all of the questions can be explained natively or via transforming to another question. While $Q_3$ is the only question with no native explanation, $Q_2$ and $Q_5$ have one dedicated technique, and the rest of the questions have multiple supporting techniques. Observe that many of the questions can be reduced to explaining unsatisfiability. Most of the methods can explain both truth values, and there is a balanced view of solver-dependency. We see quite a diversity in the explanation outputs.

\subsubsection{Other Related Explanation Methods}

The system \texttt{fasp} \cite{fichte2022rushing} allows for navigation toward desired partial solutions (called \emph{facets}) in ASP, which recently was extended to navigate the solution space of ASP-encoded planning tasks~\cite{gnad2025interactive}. 
Facets are atoms that occur in some but not all answer sets. Interactively enforcing or forbidding such facets to see which properties hold allows the user to better understand the solution space. A further recent answer set explorer is \emph{viasp}~\cite{viasp25}.

\emph{Model reconciliation} 
is a form of contrastive explanation by reconciling the system's and the 
user's model (i.e., programs) to agree on a property of the made decision, which was also considered in ASP with a focus on planning~\cite{son2021model,DBLP:conf/kr/NguyenSS020}. 

\section{Outlook for Explainable ASP}

Having reviewed the existing ASP explanation approaches, we spot gaps and present issues for future research.  

\subsection{Practical Aspects}

Notably, there are more explanation concepts than actual software tools implementing them. Issues with the tools include the following ones.


\paragraph{Language Features.} 
As shown in Section~\ref{sec:asp}, disjunction and language extensions can help with representing challenging real-world problems. 
Table~\ref{tab:tools}
summarizes the language support of available explanation tools. Although, with recent advances, there are more tools supporting language extensions, there are still some gaps.
Disjunction is often not supported by the tools, because the underlying explanation method cannot generally handle it. The reason for that being the increase in expressiveness when one adds disjunctive rules with head-cycles.
Support for aggregates and choice rules has increased, but most tools still do not support them or have some syntactic restriction on them.
Weak constraints are still generally not supported. Here DWASP is the exception, though this is a debugging tool - supporting weak constraints is independent from a diagnosis on how to restore consistency.

\paragraph{System Divide.}
The methods implemented in current systems either focus on providing local explanations or global ones. This puts the burden of deciding on which system(s) to employ on the user,  depending on the questions they have and may cause undesired overhead. 


\begin{table}[t!]
\centering
\small
\resizebox{\columnwidth}{!}{
\begin{tabular}{l l l l l l }
\toprule
\textbf{Tool} &
\textbf{Disjunction} &
\textbf{Choice rules}&
\textbf{Weak const.} &
\textbf{Aggr.}\\
\midrule
LABAS  & no & no
 & no
 & no
\\

s(CASP) & no & no
 & no
 & no
\\

UCOREX. & no & yes
 & no
 & no
\\

xclingo &yes&yes
 &no
 &no
\\


xASP2 & no &yes
 &no
 & yes\\
spock & no & no
 & no
 & no
\\

Ouroboros & yes & yes
 & no
 & no
\\

DWASP & yes & yes
 & yes  & yes
\\

\bottomrule
\end{tabular}
}
\caption{Language support of existing tools in Explainable ASP}
\label{tab:tools}
\end{table}

\paragraph{Scalability/Non-ground ASP.}

Since most of these approaches can be considered as \emph{post-hoc}, meaning that they are for explaining an encountered outcome, e.g., an answer set or none, the techniques  need to reconstruct the reasoning that achieves the result in order to explain it. This of course causes issues especially if the problem at hand is large or difficult to solve. Given that providing explanations has high computational complexity in general~\cite{DBLP:conf/jelia/EiterGO23,wang2022}, it is important to find ways to tackle scalability, e.g., by incorporating explanation techniques into the ASP solvers.
Also most explanation approaches operate on ground programs where all variables have been instantiated, except for s(CASP), which also justifies the instantiation of the variables. 
While ASP solvers 
also often only operate on ground programs, 
in the scope of explanation, preserving the relationship between ground atoms that were obtained from instantiating the same predicate may be beneficial. 

\paragraph{Complex User Questions.} Currently, in case of questions that arise over a collection of answer sets, the only way to address them is to get explanations for each answer set. However, when a user asks \emph{$Q_7$: Why is $a$ (not) in answer sets $I_1,\dots,I_n$?}, they expect a global explanation on the behaviour of the program that obtains $a$ in those particular answer sets, which currently cannot be obtained.  Explanations with respect to the quantity of answer sets, which may be needed for industrial settings with configuration problems etc., also cannot be addressed through any of the approaches. These questions can be of form \emph{$Q_8:$  Why does $P$ have so many/few answer sets?} and likely need dedicated solutions. Facets~\cite{fichte2022rushing} could be a worthwhile technique when constructing such explanations. 
Observe that these questions can also be seen as contrastive, as the user has encountered a number of answer sets against their expectation. Thus explanations on  under-/over-restrictions would be of use. Furthermore, as \emph{symmetry} is another cause for many answers, explanations that also show the detected symmetries might be helpful.


\subsection{Cognitive Aspects}

Here, we consider cognitive aspects of explanations in ASP.
 
\paragraph{Conciseness.} One challenge is reducing information complexity of explanations for large programs, 
as they may contain
many rules to trace. 
Providing more concise explanations 
from a possibly higher-level view remains open. Abstraction can be helpful to achieve this, as considered in other domains, e.g., planning \cite{sreedharan2021using} and 
mathematical reasoning \cite{poesia2022left}. 
Recent work reports a dual effect of applying abstraction in ASP explanations: abstraction by clustering helps in understanding, while
omission reduces cognitive effort~\cite{DBLP:journals/corr/abs-2602-03467}. 
These results support the need for further investigations on obtaining concise explanations. 

 \paragraph{Interpretability.} 

An important aspect when providing explanations is whether the provided explanation is indeed understandable to the user, especially if they are not experts. While justifications or similar techniques aim to provide the reasoning behind the made decision, contrastive explanations have insight into the epistemic state of the explainee and
can focus on the actual core of the misunderstanding~\cite{miller2018explanation}. 

Regarding explaining inconsistency, due its difficult nature, finding a satisfying answer
is challenging.
%
%
Concepts from debugging, like minimal unsatisfiable subsets or proof logging may be used, but it is unclear how 
respective rule sets should be presented to a non-expert. Notably, the idea was successfully applied for ASP-based product configuration~\cite{DBLP:conf/iclp/HerudBSS22}, which is a domain where minimal unsatisfiable subsets can easily be translated into an explanation.
Step-wise explanation notions \cite{bogaerts2020step,DBLP:journals/tplp/OetschPT18} could be beneficial for understanding how unsatisfiability in ASP
is brought about.

 Taking into account the aim of conciseness, 
 one idea for explaining inconsistency is by some hidden ``lemma''.
  For example, a graph colouring instance may be inconsistent due to some clique in the graph. Revealing this clique would be an adequate explanation for users with sufficient domain knowledge.
However, how such explanation could be derived is unknown. Abstraction could 
be useful but also investigating interpolation, which could produce the lemma, might be worthwhile~\cite{DBLP:journals/corr/abs-1012-3947}.


\paragraph{Interaction and Communication.}
Another issue with the available explanation techniques is the lack of user-friendly tools that allow one to obtain explanations without much technical expertise. 
In order to communicate the highly technical explanation to the end-user, Large Language Models (LLMs) might be of help. This has been explored in preliminary work 
on explanations for guess-and-check programs where questions are parsed by an LLM and the resulting MUS is translated by the LLM into natural language~\cite{geibinger2024taasp}.

Recently 
LLMs and symbolic solvers were bridged through the Model Context Protocol (MCP) \cite{szeider2025bridging}. It enables LLMs to access formal reasoning capabilities including ASP, showing potential in achieving a user-friendly interaction for ASP explanations. 
Furthermore, it remains to be explored whether LLMs can be used to find explanations for ASP.

\section{Conclusion}

We have provided an overview of explanations in ASP from an XAI perspective, structured along explanation types and on what questions the ASP user might find themselves with. We profiled a number of existing explanation methods and existing systems that can be used off-the-shelf to address the user questions. Overall, there is a rich landscape of methods and systems available with diversity in explanation types, question support, and explanation formats, but none fits all needs.

We further presented issues from practical and cognitive perspectives to address, some of which concern limitation in language support, complex user questions, and, importantly, interpretability and communication. Notably, most of the user questions can be transformed into unsatisfiability; thus  advanced explanations for 
program inconsistency
could provide a stepping stone to more universal explanation approaches. 
That and connecting to latest AI research poses a challenge.


\appendix



\section*{Acknowledgments}

This research was funded in whole or in part by the Austrian Science Fund (FWF)  grants 10.55776/COE12 and 10.55776/T1315, the Vienna Science and Technology Fund (WWTF) grant
10.47379/ICT25044, and benefited from the European Union’s Horizon
2020 research and innovation programme under grant agreement
No 101034440 (LogiCS@TUWien). \raisebox{-0.15cm}{\includegraphics[width=0.7cm]{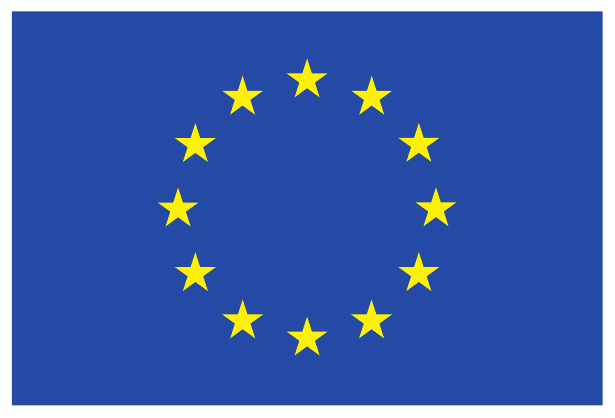}}

Tobias Geibinger is a recipient of a DOC Fellowship of the Austrian Academy of Sciences. 


\bibliographystyle{named}
\bibliography{references}

\end{document}